\documentclass[conference]{IEEEtran}
\IEEEoverridecommandlockouts
\usepackage{cite}
\usepackage{amsmath,amssymb,amsfonts}
\usepackage{algorithmic}
\usepackage{graphicx}
\usepackage{textcomp}
\usepackage{xcolor}
\usepackage{booktabs}
\usepackage{multirow}
\usepackage{threeparttable}

\def\BibTeX{{\rm B\kern-.05em{\sc i\kern-.025em b}\kern-.08em
    T\kern-.1667em\lower.7ex\hbox{E}\kern-.125emX}}
\begin{document}

\title{{\LARGE Submitted to IJCNN 2020}\\Low-Complexity LSTM Training and Inference with FloatSD8 Weight Representation}

\author{\IEEEauthorblockN{Yu-Tung Liu}
\IEEEauthorblockA{\textit{Graduate Institute of Electronics Engineering} \\
\textit{National Taiwan University}\\
Taipei, Taiwan \\
yutung860706@gmail.com}
\and
\IEEEauthorblockN{Tzi-Dar Chiueh}
\IEEEauthorblockA{\textit{Graduate Institute of Electronics Engineering} \\
\textit{National Taiwan University}\\
Taipei, Taiwan \\
chiueh@ntu.edu.tw}
}

\maketitle

\begin{abstract}
The FloatSD technology has been shown to have excellent performance on low-complexity convolutional neural networks (CNNs) training and inference. In this paper, we applied FloatSD to recurrent neural networks (RNNs), specifically long short-term memory (LSTM). In addition to FloatSD weight representation, we quantized the gradients and activations in model training to 8 bits. Moreover, the arithmetic precision for accumulations and the master copy of weights were reduced from 32 bits to 16 bits. We demonstrated that the proposed training scheme can successfully train several LSTM models from scratch, while fully preserving model accuracy. Finally, to verify the proposed method's advantage in implementation, we designed an LSTM neuron circuit and showed that it achieved significantly reduced die area and power consumption.
\end{abstract}


\section{Introduction}
Recently, a great many studies have worked on proposing low-complexity, high-performance neural networks (NN) training and inference methods as well as their hardware architectures. The main objective is to reduce the huge computational and memory storage/access needed by popular modern NN models. Research teams from academia and companies alike have presented numerous efficient acceleration solutions for convolutional NN (CNN) training and inference, based on FPGA, ASIC, and GPU. Among these methods, optimal NN model trained with single-precision floating-point (FP32) or half-precision floating-point (FP16, bfloat16) arithmetic is usually obtained and then techniques such as pruning \cite{b13}, quantization, and compression are applied. For the quantization-based precision reduction methods, recent works mostly focused on reduction of the NN inference complexity. For example, in \cite{b12}, an FP32 trained NN model was quantized into a model with fix-pointed weights that can achieve similar inference accuracy as the original FP32 NN model. 

In contrast to the many works that focused on reduced-precision NN inference, there have been several other studies that worked on reduced-precision NN training. For instance, MPT \cite{b8} trains CNNs using FP16 arithmetic and additionally proposed a loss-scaling method to preserve small gradients. In \cite{b14}, precision for weights and activations were scaled down to 4 bits with a small loss of accuracy. DoReFa-Net \cite{b15} trained NNs with 1-bit weights and 2-bit activations, while reducing the gradient precision down to 6 bits. FloatSD \cite{b3} trained CNNs with special floating-point weights with only two nonzero digits, along with 8-bit activations and gradients, and achieve very small degradation in the accuracy performance of trained NN models. Finally, \cite{b6} trains CNNs with 8-bit floating-point numbers (FP8) and proposes chunk-based accumulation and floating-point stochastic rounding to reduce the arithmetic precision for additions from 32 bits to 16 bits.

All the above proposed methods face certain problems, such as most applications were demonstrated by a few selected models, specifically only on CNNs; some methods suffer significant performance degradation; and some methods may require FP32 accumulation.

The main objective of this paper is to extend the application of reduced-precision methods to recurrent NNs (RNNs) while mitigating the aforementioned problems. FloatSD is a reduced-precision CNN training method that has been demonstrated to provide efficient training for CNNs as large as Resnet-50 \cite{b3}. Moreover, we chose long short-term memories (LSTMs) \cite{b10} as the target for low-complexity training by FloatSD. Unlike CNNs, LSTMs do not adopt convolutional layers and require only fully-connected (FC) layers consisting of simple matrix multiplications. In addition, the gradient vanishing and gradient exploding problem make LSTM very sensitive to quantization errors. These two features make the reduced-precision training of LSTMs more challenging than that of CNNs. 

We first modified the training equations and weight updates procedure of LSTM so that they become compatible with the FloatSD method. Then, a suite of LSTM models and datasets were used to demonstrate the applicability of the FloatSD technique to efficient LSTM training with no degradation in trained model accuracy. Finally, we also designed and validated a FloatSD processing element circuit suitable of low-power LSTM training acceleration.

To sum up, the contributions of this paper are:
\begin{itemize}
\item Adaptation of the FloatSD method for LSTM training.
\item Modification of the LSTM equations for lower computation complexity.
\item Precision settings for input layers, output layers, and master copy weights.
\item A low-complexity LSTM training scheme integrating the above techniques.
\item An efficient LSTM inference acceleration hardware.
\end{itemize}

\begin{figure}[b]
\centerline{\includegraphics[width=8cm]{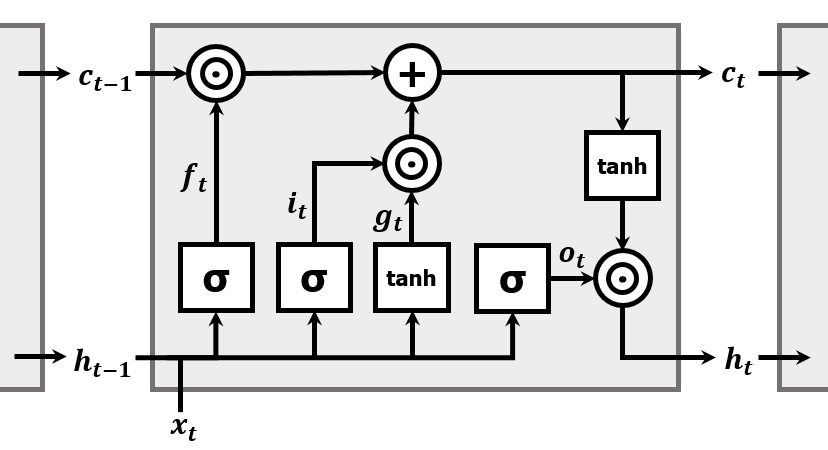}}
\caption{Architecture of an LSTM neuron.}
\label{flow}
\end{figure}

\section{Background}
\subsection{Long Short-Term Memory}

LSTM is one of the most popular types of RNNs. It can perform very well in dealing with sequence related tasks and is widely used in areas like speech recognition and natural language processing (NLP). 
As traditional RNNs, LSTM takes sequence input $x_{t}$ and generates sequence output $h_{t}$, which is computed by the following equations:
\begin{equation} \label{eq:1}
f_{t} = \sigma(W_{fx}x_{t} + W_{fh}h_{t-1} + b_{f}),
\end{equation}
\begin{equation} \label{eq:2}
i_{t} = \sigma(W_{ix}x_{t} + W_{ih}h_{t-1} + b_{i}),
\end{equation}
\begin{equation} \label{eq:3}
o_{t} = \sigma(W_{ox}x_{t} + W_{oh}h_{t-1} + b_{o}),
\end{equation}
\begin{equation} \label{eq:4}
g_{t} = tanh(W_{gx}x_{t} + W_{gh}h_{t-1} + b_{g}),
\end{equation}
\begin{equation} \label{eq:5}
c_{t} = f_{t} \odot c_{t-1} + i_{t} \odot g_{t},
\end{equation}
\begin{equation} \label{eq:6}
h_{t} = o_{t} \odot tanh(c_{t}),
\end{equation}
where the forget gate $f_{t}$ selects the information to be removed from the neuron's memory at time step $t$; the input gate $i_{t}$ selects the information to be written into the neuron's memory at time step $t$; the output gate $o_{t}$ controls the amount of information stored in cell state contributing to the output; the cell gate $g_{t}$ represents the new information at time step $t$; the cell state $c_{t}$ represents the content of the neuron's memory; and $h_{t}$ represents the output of the cell at time step $t$. It is also fed back to neuron at time step $t+1$. $W$ and $b$ represent the weights and biases, respectively. Finally, $\sigma$ and $\odot$ denote the sigmoid function and the element-wise product, respectively. The architecture of an LSTM neuron is shown in Fig. 1.

Despite excellent performance in certain areas, its complex neuron processing makes LSTM more complicated than CNNs and traditional RNNs, requiring substantially more memory and computation loading. In addition, LSTM is fully-connected and therefore its training may pose a problem of large memory IO bandwidth requirement.

\begin{figure}[b]
\centerline{\includegraphics[width=\columnwidth]{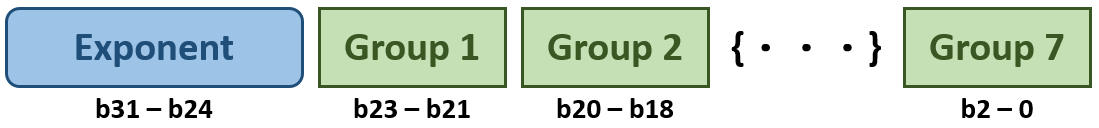}}
\caption{Structure of the FloatSD representation. This example has an 8-bit exponent field and eight three-digit groups as mantissa.}
\label{f_sd_struct}
\end{figure}

\begin{table}[b]
\caption{Proposed signed digit numbers in a 3-digit group.}
\label{sd-table}
\begin{center}
\begin{small}
\begin{tabular}{cc}
\toprule
\multirow{3}{*}{Positive}&$100 (+4)$\\\cmidrule{2-2}&$010 (+2)$\\\cmidrule{2-2}&$001 (+1)$\\ 
\midrule
\multirow{1}{*}{Zero}&$000 (0)$\\
\midrule
\multirow{3}{*}{Negative}&$00\underline{1} (-1)$\\\cmidrule{2-2}&$0\underline{1}0 (-2)$\\\cmidrule{2-2}&$\underline{1}00 (-4)$\\
\bottomrule
\end{tabular}
\end{small}
\end{center}
\end{table}

\subsection{FloatSD Number Representation}
Floating-point signed digit (FloatSD) \cite{b3} was designed for weights representation in low-complexity CNN training and inference. The structure of the FloatSD representation is shown in Fig.~\ref{f_sd_struct}. It consists of several signed digit (SD) groups and an exponent field. It is known that the complexity of multiplying two numbers depends on the number of non-zero digits in the multiplier. To achieve low multiplication complexity, each SD group allows no more than one non-zero digit. For $K$-digit SD group, there are $2K+1$ possible values. For example, in the three-digit group case, each group assume one of the seven possible values: $+4 (100)$ , $+2 (010)$, $+1 (001)$, $0 (000)$, $-1 (00\underline{1})$, $-2 (0\underline{1}0)$, $-4 (\underline{1}00)$,  as shown in Table~\ref{sd-table}. That is, the probability of a digit in an SD number with $K$-digit groups being 0 is 
$$\frac{(2K+1)K-2K}{(2K+1)K}=\frac{2K-1}{2K+1}$$
In the case of K = 3, this probability is $71.4\%$, which is even higher than the case of Canonical Signed Digit (CSD) representation (about $66.6\%$). This of course leads to the relative lower multiplication complexity in the FloatSD number representation.

\begin{figure}[t]
\centerline{\includegraphics[width=\columnwidth]{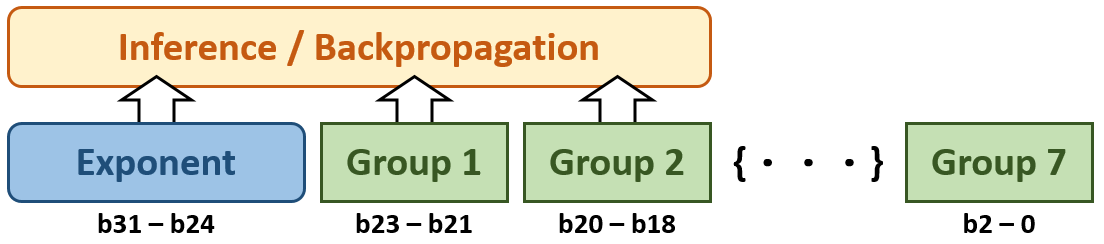}}
\caption{Using only two mantissa digit groups for inferencing and back-propagation.}
\label{f_sd_inf}
\end{figure}

Note that although having a low number of non-zero digits, FloatSD representation cannot cover all possible binary numbers. For example, a 3-digit group can only represent seven instead of eight different values by three bits in the binary representation. However, neural networks are known to be tolerant to numerical inaccuracy and sometimes even benefit from such, e.g., deliberate noise injection \cite{b9}.
To further reduce the complexity, one can adopt only a few, not all, digit groups of the FloatSD weights for inference and backward propagation, as shown in Fig.~\ref{f_sd_inf}.

\section{Low-Complexity LSTM Training}
\subsection{Weight Representation}
In this work, we chose FloatSD8 number format for LSTM weight representation. FloatSD8 consists of a 3-bit exponent field, 3-digit most significant group (MSG), and a 2-digit second group. This representation has 7 possible values ($+4$, $+2$, $+1$, $0$, $-1$, $-2$, $-4$) in the MSG and 5 possible values ($+2$, $+1$, $0$, $-1$, $-2$) in the second group, leading to 35 combinations. However, out of the 35 combinations, only 31 distinct combinations exist, making 5 bits  enough for encoding these two groups. With a 3-bit exponent field and 5-bit SD group mantissa, the FloatSD8 format requires only 8 bits for representing a neural network weight.

\subsection{Weight Update Mechanism and Master Copy}
Instead of using the FloatSD weight master copy and the Single Trigger Update (STU) method \cite{b3}, we stored the master copy of the LSTM weights during training in the conventional floating-point (FP) number format and adopted the traditional weight update mechanism. After updating, the master copy weights are then quantized to FloatSD8 for the next iteration. The change in the format of the master copy and update mechanism allows us to easily control the master copy precision without modifying the FloatSD8 format. In addition, weight initialization is very crucial to the final outcome of neural network training. As such, by choosing the FP format for weight master copy, we can adopt common weight initialization methods without modification.

\subsection{Sigmoid Function Quantization}
\begin{figure}[t]
\centerline{\includegraphics[width=8cm]{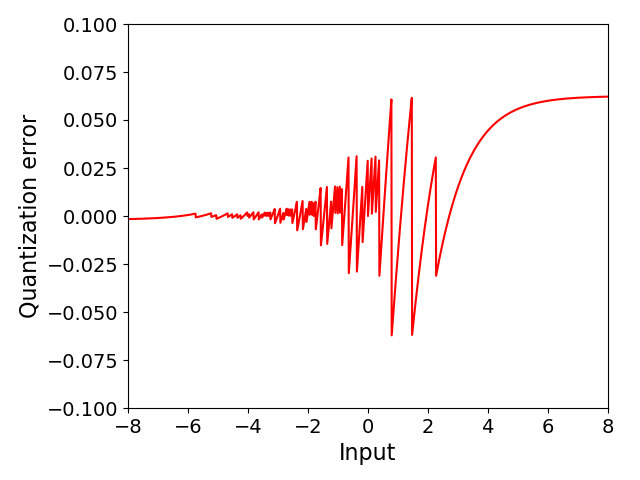}}
\caption{Quantization error of the quantized sigmoid function using only (7) for the whole input range.}
\label{f_q_error}
\end{figure}

In \eqref{eq:1}$ - $\eqref{eq:4}, the multiplications are computed between FloatSD8 weights and FP inputs, which are really efficient since a FloatSD8 weight generates only two partial products. However, in \eqref{eq:5} and \eqref{eq:6}, the element-wise multiplications are computed between two FP numbers, which would be inefficient as the FP numbers generally involve quite a few partial products. To this end, the forget gate $f_t$ , the input gate $i_t$, and the output gate $o_t$ are further quantized to the FloatSD8 representation. Then, we can convert the multiplications in \eqref{eq:5} and \eqref{eq:6} to multiplications between a FloatSD8 number and an FP number, the same as the format in \eqref{eq:1}$ - $\eqref{eq:4}.  Direct FloatSD8 quantization of the sigmoid function leads to unbalanced quantization error distribution between positive and negative inputs, as shown in Fig.~\ref{f_q_error}. This is caused by the logarithmic linear nature of the FloatSD representation.

\begin{figure}[b]
\centerline{\includegraphics[width=8cm]{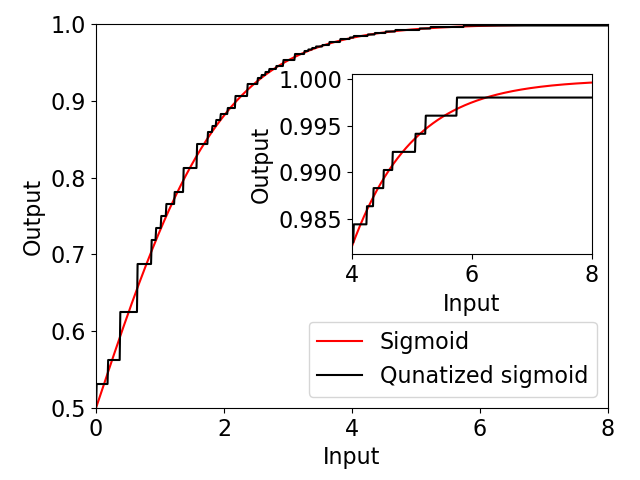}}
\caption{Sigmoid function and quantized sigmoid function (Eq. (8)) for the input range 0 to 8.}
\label{f_sigmoid}
\end{figure}

Therefore, we decompose the quantization operation into two regions by
\begin{equation}
y=Q(\sigma(x)),\ \text{for}\ x\leq0,
\end{equation}
\begin{equation} 
y=1-Q(\sigma(-x)),\ \text{for}\ x>0,
\end{equation}
where $Q(\cdot)$ denotes the FloatSD8 quantization function. The quantized sigmoid function and the un-quantized counterpart for $0 < x < 8$ are plotted in Fig.~\ref{f_sigmoid}. Note that in Eq. (8), the output $y$ may need two FloatSD8 numbers to represent. 

In actual implementation, the sigmoid function and the FloatSD quantization can be merged and realized by a lookup table (LUT). The extra multiplications and additions from two FloatSD numbers representing one quantized sigmoid output can be handled by the specially designed multiply and accumulate (MAC) circuit. Moreover, because there are only 42 possible values in a quantized sigmoid output when the input is non-positive, the depth of the LUT can be reduced, significantly lowering the memory requirement. 

\subsection{Other Low-Complexity Considerations}
The FloatSD8 representation is used for the LSTM network weights. However, LSTM training and inference computation involves more than just network weights. In this work, forward neuron activations, backward neuron activations, and all gradients were quantized to 8-bit FP number (FP8), having 1-bit sign, 5-bit exponent, and 2-bit mantissa \cite{b6}. Note that although quantization by stochastic rounding is shown to provide better training performance, all the above quantization adopted the regular rounding in consideration of hardware design complexity.

\subsection{Summary}
The precision settings of the proposed low-complexity LSTM training scheme are summarized in Table~\ref{t_precision}. By quantizing these variables, LSTM training can benefit from not only low hardware complexity but also low memory access bandwidth.

\begin{table}[b]
\caption{Precision setting of the proposed LSTM training scheme.}
\label{t_precision}
\begin{center}
\begin{small}
\begin{threeparttable}
\begin{tabular}{ccccc}
\toprule
$w^{\mathrm{a}}$ & $g^{\mathrm{b}}$ & $a^{\mathrm{c}}$ & $m^{\mathrm{d}}$ & $s^{\mathrm{e}}$ \\
\midrule
FloatSD8 & FP8 & FP8 & FP32 & FloatSD8 \\
\bottomrule
\end{tabular}
\begin{tablenotes}[para,flushleft]
  $^{\mathrm{a}}$Weights, $^{\mathrm{b}}$Gradients,  $^{\mathrm{c}}$Activations, $^{\mathrm{d}}$Master copy of weights, $^{\mathrm{e}}$Quantized sigmoid function output
 \end{tablenotes}
\end{threeparttable}
\end{small}
\end{center}
\end{table}

\section{Simulation and Discussion}
\subsection{Platform, Datasets, and Models}
PyTorch and QPyTorch \cite{b7} were used as the frameworks to study the proposed method. Four commonly used NLP datasets – UDPOS \cite{b5}, SNLI \cite{b1}, Multi30K \cite{b2}, and WikiText-2 \cite{b4} were used in the simulations. All networks are trained via the proposed FloatSD8 training method with the same network architectures, hyperparameters, and other preprocessing as the baseline implementation using the IEEE single-precision floating-point (FP32) arithmetic. One exception is the loss-scaling technique \cite{b8} was adopted in the low-complexity training method to limit the back-propagated error magnitude within a small interval. In the following, we briefly introduce each dataset, the corresponding model, and the hyperparameters used.
\paragraph{UDPOS}
UDPOS comprises 254,830 words and 16,622 sentences taken from five genres of web media, with sentences annotated using universal dependency relations. For UDPOS simulation, we adopted a model consisting of a word embedding layer, two-layer bidirectional LSTM, and a fully-connected output layer. The model was trained via the ADAM \cite{b11} optimizer with a single scaling factor of 1024.

\paragraph{SNLI}
Stanford Natural Language Inference (SNLI) dataset is a collection of 570k human-written English sentence pairs, which are manually labeled for balanced classification with the natural language inference (NLI) labels that are either entailment, contradiction, or neutral. For SNLI simulation, we adopted a model that consists of a word embedding layer, a fully connected projection layer, a single-layer bidirectional LSTM, and a sequence of four fully-connected layers. The model was trained via the ADAM optimizer with a single scaling factor of 1024.

\paragraph{Multi30K}
Multi30K consists of 29,000 training data and 1,014 development data, each containing an English source sentence, German translation by humans, and a corresponding image. The dataset is for the multimodal translation task that translates the English sentence describing an image into German. For Multi30K simulation, we adopted the model with an encoder and a decoder. The encoder is made up of a word embedding layer and a single-layer LSTM; and the decoder consists of a word embedding layer, a single-layer LSTM, and a fully-connected output layer. The model was trained via the ADAM optimizer with a single scaling factor of 1024.

\paragraph{WikiText-2}
The WikiText language modeling dataset (WikiText-2) is a collection of over 100 million tokens extracted from articles in Wikipedia. This dataset is relatively bigger than the previous three datasets, providing us insights into the effectiveness of the proposed training scheme on a huge dataset. For WikiText-2 simulation, we adopted a model with a word embedding layer, a two-layer LSTM encoder, and a fully-connected output decoder. The model was trained via the SGD optimizer with a single scaling factor of 1024.

\begin{table}[b]
\caption{Hyperparameter settings and parameter counts in four datasets.}
\label{t_hyper}
\begin{center}
\begin{small}
\begin{tabular}{cccc}
\toprule
Dataset & Epoch & Batchsize & Parameters \\
\midrule
UDPOS    &50&64&0.64M\\
SNLI    &30&128&4.23M\\
Multi30K    &30&128&15.27M\\
WikiText-2    &50&64&84.98M\\
\bottomrule
\end{tabular}
\end{small}
\end{center}
\end{table}

\subsection{Simulation Results and Discussion}
\begin{figure}[t]
\centerline{\includegraphics[width=\columnwidth]{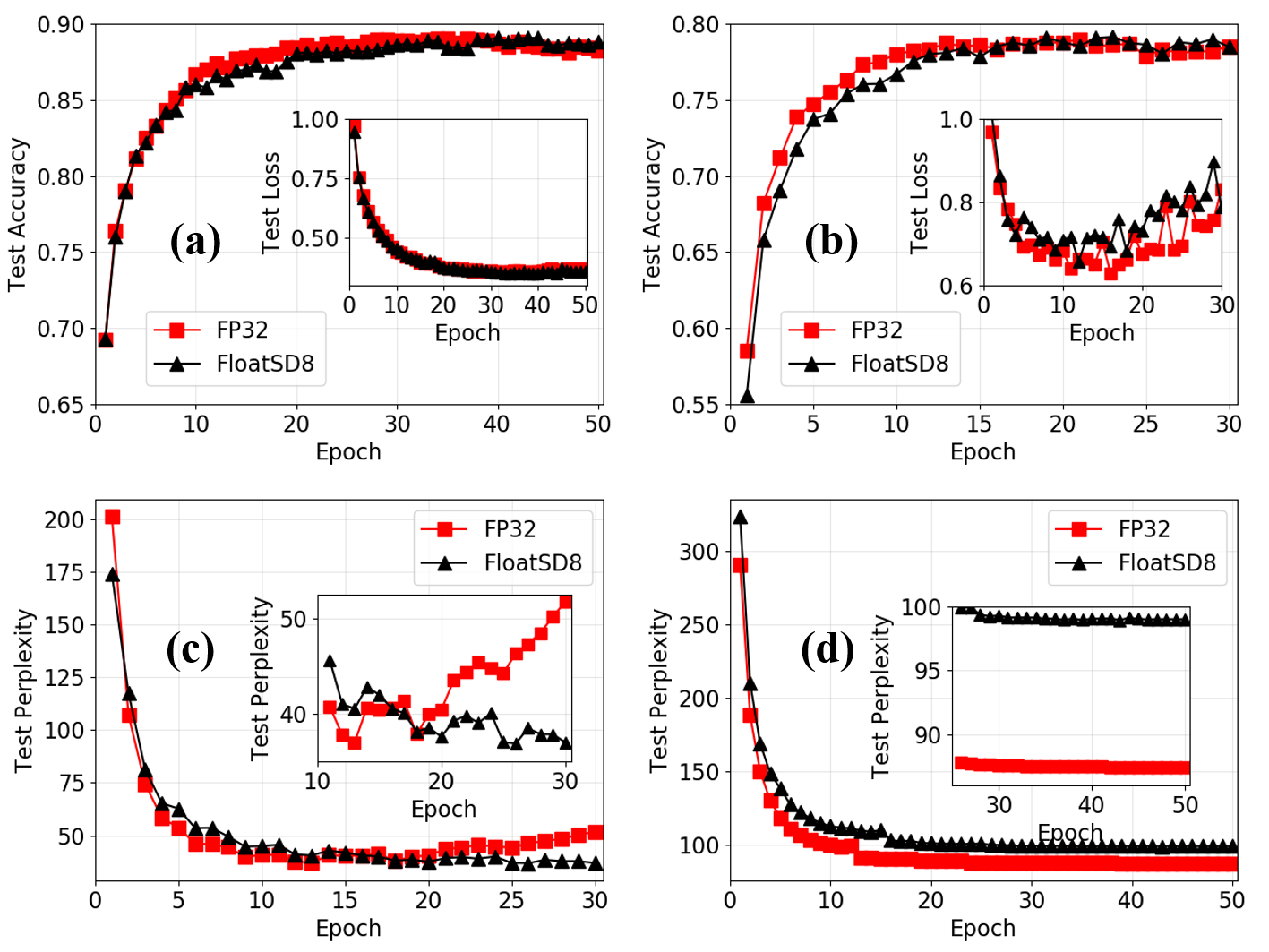}}
\caption{Simulated performance curves of the proposed FloatSD8 training scheme and the FP32 baseline. (a) UDPOS, (b) SNLI, (c) Multi30K, (d) WikiText-2.}
\label{f_result_first}
\end{figure}

The hyperparameter settings and parameter counts of these four datasets are summarized in Table~\ref{t_hyper}. The simulated performance curves during FloatSD8 training and FP32 training of the four datasets are shown in Fig.~\ref{f_result_first}. It is clear that when compared with the FP32 trained model the proposed FloatSD8 training scheme can achieve similar or even better performance in UDPOS, SNLI, and Multi30K applications. However, in the WikiText-2 task, degradation in perplexity by the proposed method is quite obvious. The simulated FloatSD8/FP32 trained LSTM results are summarized in the second and third columns in Table~\ref{t_result_summary}.

\paragraph{Activation Precision of the First and Last Layers}
In NN training, the first and last NN layers are often excluded from quantization due to their sensitivity. Previous simulations quantized the forward and backward activations of the first and the last layers in FP8. This may make no difference in relatively small datasets, however, in the larger dataset as WikiText-2, the poor precision in these two layers may significantly impact the training performance. To gain more insight into the cause of the performance degradation in WikiText-2 task, we experimented with various settings of activations using the same model architecture and hyperparameters. The results are summarized in Table~\ref{t_result_wiki}. Note that the first layer in the table means the outputs of the embedding layer since the inputs of the embedding layer are just indices. From the results, we can conclude that the last layer’s activation precision is more important than the first layer’s activation precision. Also, the setting of using FP8 first layer activations, FP16 last layer activations, and FP8 other layers activations are sufficient to provide similar results comparing to the FP32 baseline. Note that this way we relaxed only the output layer activation precision to FP16, while keeping the weights and all other activations in 8-bit precision. All the multiplications in the LSTM were still between FloatSD8 and FP8, except that the output layer activations were not further quantized to FP8.

\begin{table}[t]
\caption{The simulation results across four datasets.}
\label{t_result_summary}
\begin{center}
\begin{small}
\begin{threeparttable}
\begin{tabular}{cccc}
\toprule
Dataset &  FP32 baseline & FloatSD8 & FloatSD8$^{\mathrm{e}}$ \\
\midrule
UDPOS$^{\mathrm{a}}$    & 89.05 & 89.09 & 89.13\\
SNLI$^{\mathrm{b}}$    & 79.28 & 79.32 & 79.24\\
Multi30K$^{\mathrm{c}}$    & 37.02 & 36.87 & 37.26\\
WikiText-2$^{\mathrm{d}}$   & 87.83 & 98.94 & 91.06\\
\bottomrule
\end{tabular}
\begin{tablenotes}[para,flushleft]
$^{\mathrm{a}}$Accuracy(\%),  $^{\mathrm{b}}$Accuracy(\%), $^{\mathrm{c}}$Perplexity, $^{\mathrm{d}}$Perplexity $^{\mathrm{e}}$With FP16 master copy of weights
\end{tablenotes}
\end{threeparttable}
\end{small}
\end{center}
\end{table}

\begin{table}[t]
\caption{Simulated perplexity results of the WikiText-2 dataset using five different activation precision settings.}
\label{t_result_wiki}
\begin{center}
\begin{small}
\begin{tabular}{cccc}
\toprule
First layer& Last layer & Other layers & Perplexity \\
\midrule
FP8 & FP8 & FP8 & 98.94 \\
FP16 & FP16 & FP16 & 88.92 \\
FP8 & FP16 & FP8 & 89.87 \\
FP16 & FP8 & FP8 & 99.81 \\
FP16 & FP16 & FP8 & 89.59 \\
\bottomrule
\end{tabular}
\end{small}
\end{center}
\end{table}

\paragraph{Precision of the Master Copy Weights}
The master copy weights used in the previous experiments were in the FP32 format. If we can reduce the precision of the master copy, both memory and complexity can be saved. As such, we further experimented with the four datasets using the FP16 master copy weights during training. Note that the simulations were done without any other change in model architecture or hyperparameters, except changing the FP32 master copy to FP16 precision. Simulated results of the four datasets are summarized in the fourth column in Table~\ref{t_result_summary}. Comparing to their FP32 counterparts, the results using the FP16 master copy have quite similar performance. The highest degradation between using all FP32 arithmetic and FloatSD8/FP8 with FP16 output activations (see Table IV) and FP16 master copy appears on the WikiText-2 dataset, which is only about 3.7\% in degradation in perplexity.

In conclusion, the modified FloatSD8 training scheme, i.e. FloatSD8 weights, FP16 master copy, FP8 gradients, FP8 forward and backward activations, except for FP16 last layer’s outputs, and FloatSD8 sigmoid function quantization, can achieve low complexity training and inference across different LSTM models as well as comparable performance with negligible degradation when compared to the  baselines trained in FP32 arithmetic. The precision setting of the modified training scheme is summarized in Table~\ref{t_precision_modified}.

\begin{table}[b]
\caption{Precision setting of the modified LSTM training scheme.}
\label{t_precision_modified}
\begin{center}
\begin{small}
\begin{threeparttable}
\begin{tabular}{cccccc}
\toprule
w$^{\mathrm{a}}$ & g$^{\mathrm{b}}$ & o$^{\mathrm{c}}$ & a$^{\mathrm{d}}$ & m$^{\mathrm{e}}$ & s$^{\mathrm{f}}$ \\
\midrule
FloatSD8 & FP8 & FP16 & FP8 & FP16 & FloatSD8 \\
\bottomrule
\end{tabular}
\begin{tablenotes}[para,flushleft]
$^{\mathrm{a}}$Weights, $^{\mathrm{b}}$Gradients,  $^{\mathrm{c}}$Activations of the last layer output,  $^{\mathrm{d}}$Activations of other layers,  $^{\mathrm{e}}$Master copy of weights,  $^{\mathrm{f}}$Outputs of the sigmoid function, 
\end{tablenotes}
\end{threeparttable}
\end{small}
\end{center}
\end{table}

\subsection{Complexity Analysis}
FloatSD8 represents a network weight with no more than two non-zero digits and an exponent field. Consequently, multiplication involving a FloatSD8 weight can be implemented by addition of two partial products. As the neuron activations are quantized to the FP8 format, the forward pass multiplication is done by addition of two partial products generated from the FP8 multiplicand, and requires only FP16 additions. In the backward pass, FP16 additions also suffice because the backward neuron activations are also quantized to the FP8 format. The weight update process is implemented by addition of the FP16 master copy weight and the FP8 gradient, which can also be realized by FP16 addition. In conclusion, FP16 accumulation is sufficient for all operations in LSTM model training and inference.

\begin{figure}[b]
\centerline{\includegraphics[width=\columnwidth]{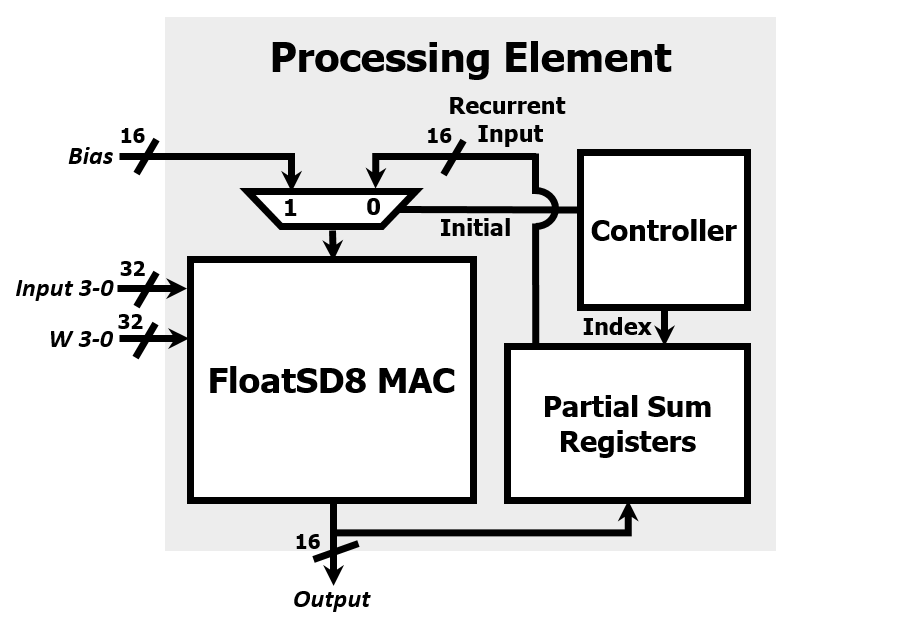}}
\caption{Architecture of the LSTM processing unit.}
\label{f_pe}
\end{figure}

\section{Hardware Implementation}
Based on the FloatSD8 weight representation, FP8 input activation, sigmoid function FloatSD8 quantization, and FP16 accumulation, we designed a low-complexity LSTM inference accelerator circuit that aims to leverage the low precision variable representation of the proposed method.

\subsection{Processing Element}
The LSTM processing elements (PE) is the core circuit of the proposed hardware accelerator. The PE computes matrix multiplication between FP8 inputs and FloatSD8 weights. The architecture of the PE is illustrated in Fig.~\ref{f_pe}. Since the input size is influenced by the varying input sequence length, the LSTM PE adopts the output-stationary design, and accumulates the product sum generated by the current batch of inputs/weights in the partial sum register. The FloatSD8 MAC takes four FP8 inputs, four FloatSD8 weights and the previous results or the bias as input data; computes multiplications between inputs and weights; and then accumulates all products and the previous result or the bias. Taking advantage of 8-bit inputs/weights, the FloatSD8 MAC simultaneously handles four pairs of inputs and weights using the same IO bandwidth as an FP32 MAC.

\begin{figure}[t]
\centerline{\includegraphics[width=\columnwidth]{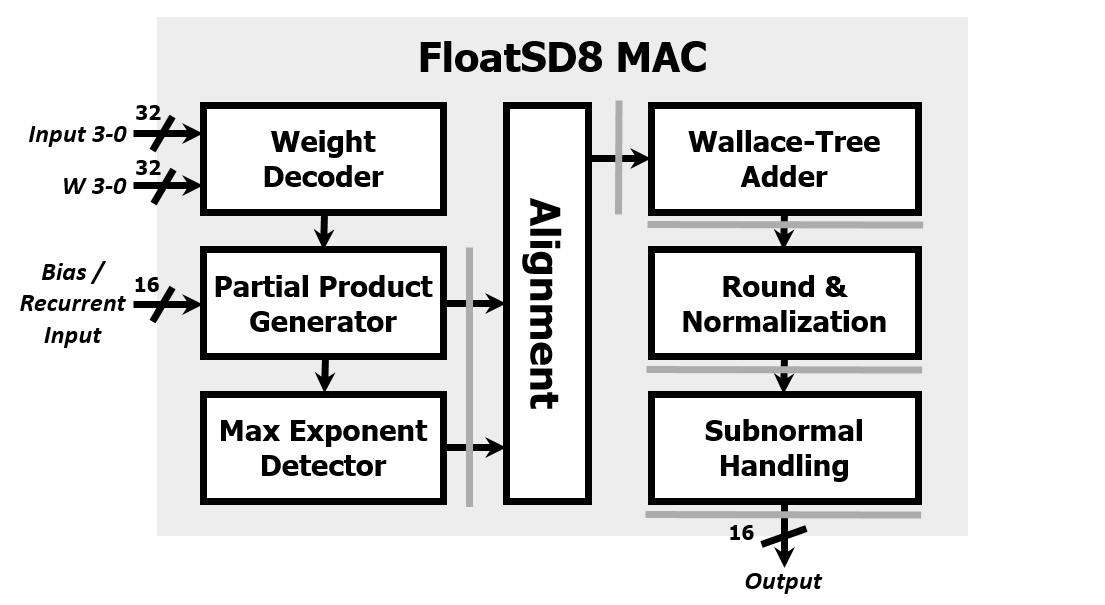}}
\caption{Block diagram of the proposed FloatSD8 MAC.}
\label{f_mac}
\end{figure}

The block diagram of the proposed five-stage pipelined FloatSD8 MAC is depicted in Fig.~\ref{f_mac}. In the first stage, FloatSD8 weights are decoded; the partial product generator then generates partial products between four pairs of inputs and weights, and the max exponent detector finds the largest exponent among all partial products. In the second stage, the partial products are aligned by respective shifters. In the third stage, aligned partial products are added by Wallace-tree type carry-save adders. Finally, in the fourth and fifth pipeline stages, the result is rounded and normalized to the FP16 format. Note that the FloatSD8 MAC takes the previous result as one input, so the PE would have to wait for five cycles before computing another outcome, leading to low throughput and low hardware utilization. To overcome this problem, batch workloads are adopted in our design with the partial sum registers. With the batch size larger than five, the hardware utilization would reach 100\%.

\begin{figure}[b]
\centerline{\includegraphics[width=\columnwidth]{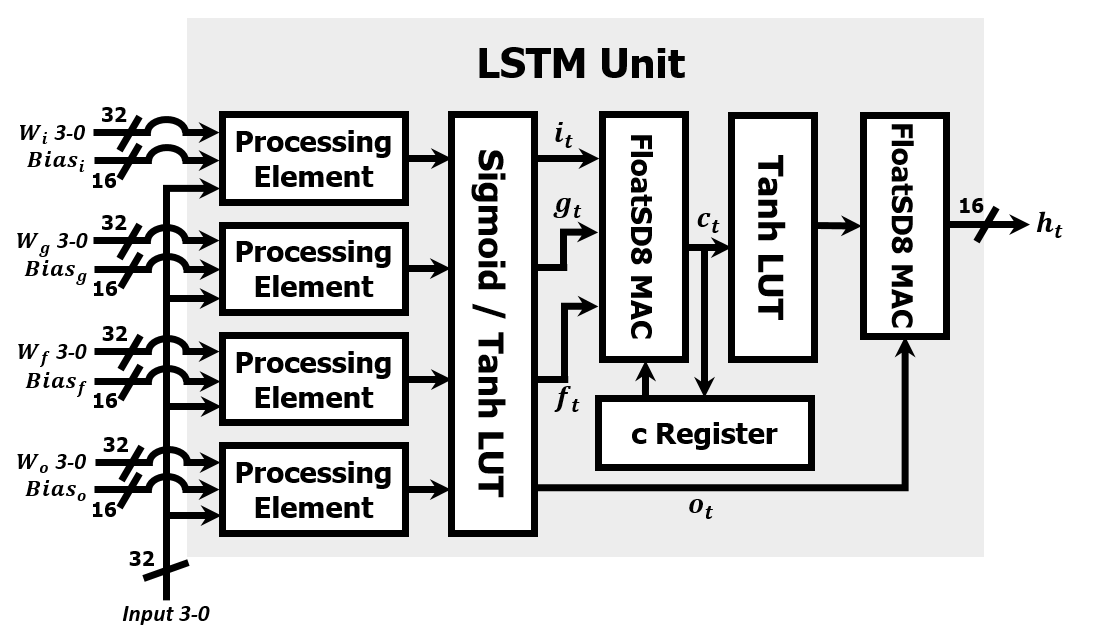}}
\caption{Block diagram of FloatSD8 based LSTM neuron circuit.}
\label{f_unit}
\end{figure}

\subsection{LSTM Unit}
The architecture of the LSTM inference circuit is shown in Fig.~\ref{f_unit}. It consists of four PEs, LUTs for sigmoid and tanh function, memory for the cell state, and two FloatSD8 MAC. For the computation of the whole LSTM operation, the inputs and weights would first be sent to four PEs for calculation of matrix multiplications in \eqref{eq:1} – \eqref{eq:4}. After completing matrix multiplications, the outputs of PEs would then be sent to LUTs, getting the results of four gates. The FloatSD8 MAC would then compute the cell state $c_t$ according to \eqref{eq:5}. As mentioned before, the outputs of sigmoid function LUT are two FloatSD8 format numbers, so the computation of the cell state $c_t$ would be multiply-accumulate operation between four FloatSD8 numbers and four FP8 numbers, exactly what a FloatSD8 MAC can handle. Finally, the cell state would be sent to a tanh LUT, and then the FloatSD8 MAC would calculate the output $h_t$ according to \eqref{eq:6}.

\subsection{Area and Power Comparison}
In order to demonstrate the effectiveness of the proposed FloatSD8 method and its associated circuit, we also designed an FP32 MAC that takes four pairs of inputs and weights as input data. The FP32 MAC was properly pipelined to run at the same speed as the FloatSD8 MAC. These two MACs were then synthesized in Synopsys Design Compiler using a 40nm CMOS process. Moreover, Synopsys PrimeTime PX was used for accurate power estimation. Table~\ref{t_hardware} lists the estimated area and power consumption of the two synthesized MAC circuits. When running at 400MHz, the FloatSD8 MAC is about 7.66X smaller in die area and consumes 5.75X less power than the FP32 MAC running at the same speed. The significant saving in circuit area and power consumption indeed validate the effectiveness of the proposed FloatSD8 design for LSTM applications.

\begin{table}[b]
\caption{Power and area comparison between MACs based on FloatSD8 and FP32.}
\label{t_hardware}
\begin{center}
\begin{small}
\begin{tabular}{ccccc}
\toprule
Process & Type & Period & Area & Power \\
\midrule
40nm CMOS & FP32 & 2.5ns & 26661${\mu}m^{2}$ & 2.920$mW$ \\
40nm CMOS & FloatSD8 & 2.5ns & 3479${\mu}m^{2}$ &0.508$mW$ \\
\bottomrule
\end{tabular}
\end{small}
\end{center}
\end{table}

\section{Conclusions}
In this paper, we applied the novel FloatSD8 weight representation to LSTM training and inference. To fully leverage the low complexity feature of the FloatSD-based multiplier, the sigmoid function in LSTM operation is cascaded with FloatSD8 quantization. With this modification, we can execute all multiplications in  the LSTM network by an FP8-FloatSD8 multiplier using only two partial products. To further reduce the computational cost and storage plus IO access cost in LSTM training, we used 8-bit FP8 gradients and activations, 16-bit accumulation, and the master copy weights. Simulation of four different LSTM applications indicate that the proposed FloatSD8 based training method can achieve almost the same and in some cases better performance when compared to FP32 baselines. To more convincingly verify the advantage of the proposed method, we designed an LSTM inference acceleration circuit for the proposed FloatSD technology. We show that our design outperforms the LSTM circuit based on FP32 arithmetic in both die area and power consumption, respectively by 7.66X and 5.75X. Finally, based on this work and the previous FloatSD work on CNNs [3], we believe that the FloatSD8 representation is suitable for NN training across different domains and model architectures.

In the future, we plan to train more types of NNs using the FloatSD technology to broaden its scope of application. Meanwhile, a general-purpose high-performance FloatSD8-based NN training/inference accelerator SoC has been taped out and will be tested soon.

\end{document}